\documentclass[runningheads]{llncs}

\usepackage{eccv}

\newif\ifSupp
\Supptrue

\usepackage{diagbox}
\usepackage{multirow, graphicx}
\usepackage{url}
\usepackage{graphicx}
\usepackage{amsmath,amsfonts,amssymb,bm}
\usepackage{tabularx}
\usepackage{tikz}
\usepackage{makecell}
\usepackage{multicol,lipsum}
\usepackage{changepage}
\usepackage{arydshln}
\usepackage{algpseudocode}
\usepackage{enumitem}
\usepackage{mathtools}
\usepackage{colortbl}

\usepackage{amsmath,amsfonts,bm}

\def\eqref#1{equation~\ref{#1}}

\def\eps{{\epsilon}}

\DeclareMathAlphabet{\mathsfit}{\encodingdefault}{\sfdefault}{m}{sl}
\SetMathAlphabet{\mathsfit}{bold}{\encodingdefault}{\sfdefault}{bx}{n}

\newcommand{\set}[1]{\left\{#1\right\}}

\newcommand{\norm}[1]{\left\Vert#1\right\Vert}
\newcommand{\abs}[1]{\left\vert#1\right\vert}

\newlength{\Oldarrayrulewidth}

\newcommand{\gc}{\cellcolor[gray]{0.9}}
\newcommand{\ul}[1]{{\underline{#1}}}

\usepackage{eccvabbrv}

\usepackage{graphicx}
\usepackage{booktabs}

\usepackage[accsupp]{axessibility}  

\usepackage[pagebackref,breaklinks,colorlinks,citecolor=eccvblue]{hyperref}

\usepackage{orcidlink}

\begin{document}

\title{Improving Adversarial Transferability via \\Model Alignment}


\author{
Avery Ma\inst{1,2}
\and
Amir-massoud Farahmand\inst{1,2} 
\and 
Yangchen Pan\inst{3}
\and \\
Philip Torr\inst{3}
\and
Jindong Gu\inst{3}
}

\authorrunning{A.~Ma et al.}

\institute{University of Toronto \and Vector Institute \and University of Oxford
}

\maketitle
\begin{abstract}
Neural networks are susceptible to adversarial perturbations that are transferable across different models. In this paper, we introduce a novel model alignment technique aimed at improving a given source model's ability in generating transferable adversarial perturbations. During the alignment process, the parameters of the source model are fine-tuned to minimize an alignment loss. This loss measures the divergence in the predictions between the source model and another, independently trained model, referred to as the witness model. To understand the effect of model alignment, we conduct a geometric analysis of the resulting changes in the loss landscape. Extensive experiments on the ImageNet dataset, using a variety of model architectures, demonstrate that perturbations generated from aligned source models exhibit significantly higher transferability than those from the original source model. Our source code is available at \mbox{\url{https://github.com/averyma/model-alignment}}.
\end{abstract}



\section{Introduction}\label{sec:intro}
Adversarial examples with small perturbations can mislead deep neural networks to make wrong predictions~\cite{szegedy2014intriguing}. It has been observed that perturbations created to attack one model can also fool other models, which is known as adversarial transferability~\cite{goodfellow2014explaining}. The transferability of such perturbations has received significant attention recently since it poses practical concerns for the deployment of machine learning systems in real-world applications~\cite{qayyum2020secure,deng2020analysis,li2020practical}. 

One possible explanation for this transferability is that transferable perturbations exploit similar features present in both the source and target models~\cite{ilyas2019adversarial}. To see this, let us first consider the hypothesis that neural networks capture two distinct types of features from data: semantic features and human-imperceptible features. This hypothesis has been proposed and supported in \cite{ma2023understanding, wang2020high, yin2019fourier}. We provide a summary of their findings as follows. First, models learn semantic features that align with human perception. The extraction of such features is similar across different models, reflecting a shared understanding of the semantics. Second, models learn human-imperceptible features, and their learning is model-specific. For example, Ma \etal~\cite{ma2023understanding} demonstrated that the use of these features varies based on models’ initialization and their optimization process, while Wang \etal~\cite{wang2020high} discussed how model architecture can result in model-specific interpretations of these features.

Features and vulnerabilities are highly correlated, as vulnerabilities often stem from the exploitation of certain features~\cite{ilyas2019adversarial}. In the context of transferability, the degree of similarity between those exploited features in the source and target model is crucial. That is, the more similar the exploited features between models, the more likely it is that the perturbation will successfully transfer. To support this, Liu \etal~\cite{liu2016delving} showed that different models have similar decision boundaries (from learning similar features), thus enabling some perturbation to be transferable across different models. However, some perturbations exploit features that are specific to the source model. Qin \etal~\cite{qin2022boosting} demonstrated that when maximizing the cross entropy loss to find adversarial examples, some perturbations fail to transfer since they are located at sharp local maxima unique to the source model, which do not exist for the target model.

Motivated by these observations, we propose a model alignment technique to modify the source model to encourage a similar feature extraction as other, independently trained models, which we refer to as \textit{witness models}. 
\begin{adjustwidth}{0.8cm}{0.8cm}
    \textbf{\textit{Our goal is to transform any source model into one from which attacks generate more transferable perturbations.}}
\end{adjustwidth}
\noindent During the alignment process, the source model's parameters are fine-tuned to minimize an alignment loss, which measures the prediction divergence between the source and witness models. Through this alignment process, the source model learns to focus on a set of features that are similarly extracted by the witness model. This allows attack algorithms to more effectively exploit features common across models, leading to more transferable perturbations.
\begin{figure}[t]
    \captionsetup[subfigure]{labelformat=empty, labelsep=period}
    \centering 
    \subfloat[]{\includegraphics[width=\linewidth]{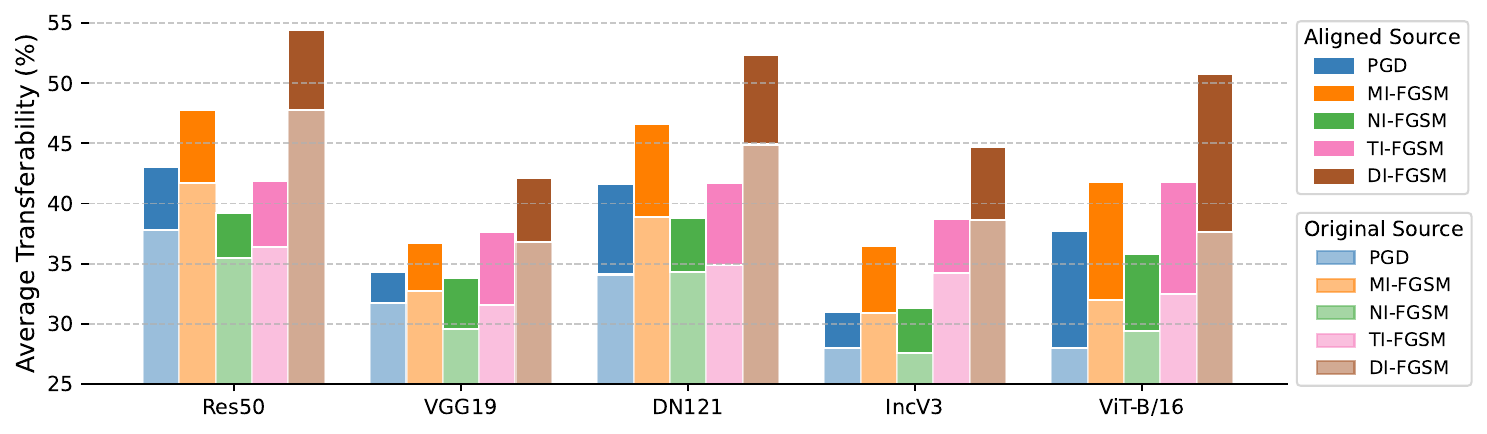}}
    %
    \caption{\textbf{Attacking the aligned source model for more transferable perturbations.}
    We compare the transferability of $\ell_\infty$-norm bounded perturbations ($\epsilon=4/255$) generated using the source model before and after performing model alignment. The result highlights the \textbf{compatibility} of model alignment with a wide range of attacks, as perturbations generated from the aligned source model become more transferable. Here, the source model is aligned using a witness model from the same architecture but is initialized and trained independently. Results are averaged over all target models.}
    \label{fig:teaser}
\end{figure}

Because of the orthogonal nature of our approach, model alignment complements, rather than competes with, other attack algorithms. This synergy underscores our method's key advantage: its broad compatibility with a wide range of attack algorithms, as highlighted in Fig.~\ref{fig:teaser}. Extensive experiments on various combinations of architectures, transferability on individual target models, and results with other attacks are included in Sec.~\ref{sec:exp}.

We analyze the effect of the proposed alignment method, demonstrating that perturbations generated from the source model exploit more semantic features that are shared across different models.
Additionally, we conduct a geometric analysis to study the changes in the loss landscape resulting from this process. The analysis explores the relationship between model alignment and the use of soft labels~\cite{szegedy2016rethinking, he2019bag, zhang2018mixup}.
Results show that there is a notable smoothing effect on the loss surface of the model, which is consistent with prior findings that adversarial examples from flatter maxima are more transferable~\cite{gubri2022lgv, qin2022boosting}. Our contributions can be summarized as follows:
\begin{itemize}
    \item We present a model alignment method to fine-tune the source model by minimizing an alignment loss which measures the difference in the output between the source model and the witness model.
    \item To understand the effect of model alignment, we conduct a geometric analysis to study the changes in the loss landscape resulting from this process.
    \item Extensive experiments on ImageNet~\cite{imagenet}, using convolutional neural networks (CNNs) and Vision Transformers (ViTs)~\cite{dosovitskiy2020image, liu2021swin}, demonstrate that perturbations generated from aligned source models exhibit significantly higher transferability than those from the original source model. We demonstrate that our alignment technique is compatible with a wide range of attacks.
\end{itemize}
\section{Related Work}
In this section, we provide a brief overview of approaches to generate more transferable perturbations. For a more comprehensive review of research related to adversarial transferability, we direct the reader to a recent survey~\cite{gu2024survey}.

\subsection{Generating Transferable Perturbations}
Existing work on improving the transferability of adversarial examples can be categorized into four groups: data-augmentation-based methods~\cite{xie2019improving, dong2019evading, zou2020improving, wu2021improving, li2020regional, byun2022improving, cheng2024typography}, optimization-based methods~\cite{dong2018boosting, lin2019nesterov, wang2021enhancing, zhao2021success, zhang2022investigating, xiao2021improving, han2023ot, luo2023image}, model-modification-based methods~\cite{benz2021batch, wu2020skip, guo2020backpropagating, xiaosen2024rethinking}, and ensemble-based methods~\cite{liu2016delving, gubri2022lgv, qian2023lea2, li2022making}.

\noindent\textbf{Data-augmentation-based methods:}
Data augmentation prevents overfitting in deep neural networks, with advanced techniques~\cite{zhang2018mixup, yun2019cutmix, cubuk2019autoaugment,ma2022sage} being key for state-of-the-art generalization on large datasets like ImageNet. Building on this concept, several works have proposed the incorporation of various data augmentation techniques into the attack algorithm. This integration aims to prevent adversarial examples from overfitting to the source model, thereby improving their transferability.

\noindent\textbf{Optimization-based methods:}
Lin \etal~\cite{lin2019nesterov} drew a parallel between generating transferable adversarial examples and training neural networks.
In this analogy, source models are the training data, adversarial perturbations are model parameters, and the target model is the testing data. Thus, transferability of adversarial examples is akin to model generalization. Optimization-based methods, such as momentum~\cite{dong2018boosting, lin2019nesterov, zou2020improving} and variance tuning~\cite{wang2021enhancing, xiong2022stochastic}, initially designed to improve model generalization, can similarly improve adversarial transferability.

\noindent\textbf{Model-modification-based methods:}
Several studies have proposed methods to improve transferability by modifying the source model. Benz \etal~\cite{benz2021batch} showed that perturbations from models without batch normalization~\cite{ioffe2015batch} are more transferable. Methods like Linear Backpropagation (LinBP)~\cite{guo2020backpropagating} and Backward Propagation Attack (BPA)~\cite{xiaosen2024rethinking} focus on non-linear activations and modify the derivative of ReLU. Wu \etal~\cite{wu2020skip} showed that increasing gradients from skip connections can improve transferability. A key advantage of our approach is its model-agnostic nature: alignment can be applied without changing the model's forward or backward pass, improving any source model's ability to generate more transferable perturbations. In contrast, other methods require changes, such as those seen in LinBP and BPA, or even complete retraining.

\noindent\textbf{Ensemble-based methods:}
Another line of approaches involves the use of multiple models for generating adversarial examples. Liu \etal~\cite{liu2016delving} were among the first to propose enhancing transferability by attacking an ensemble of models, with the rationale being that a perturbation capable of fooling multiple models is more likely to deceive the target model. More recently, Gubri \etal~\cite{gubri2022lgv} proposed constructing an ensemble of source models by collecting weights along the fine-tuning trajectory of a trained model. 

\subsection{Understanding Adversarial Transferability}
Several works have focused on understanding the transferability of adversarial perturbations~\cite{zhang2024does, wu2020towards, waseda2023closer, zhu2021rethinking}, with some analyzing from a geometric perspective~\cite{fawzi2017robustness, charles2019geometric, zhao2020bridging, liu2016delving}. Liu \etal~\cite{liu2016delving} showed that weakly transferable adversarial examples often fall into local maxima of the source model. They found that some perturbations fail to transfer because they reside in tiny pockets corresponding to the ground truth label, present in the source model but not in the target model. Similarly, Gubri \etal~\cite{gubri2022lgv} hypothesized that adversarial examples at flat loss maxima transfer more effectively than those at sharp maxima. Motivated by these insights, in Sec.~\ref{sec:geometric_analysis}, we extend this geometric perspective to examine how the alignment method influences the source model's loss surface geometry to generate more transferable adversarial examples.
\section{Method}
We present the model alignment process and understand why aligned models can generate more transferable perturbations.

\subsection{Preliminary}
In this work, we focus on neural networks used for classification tasks. Let us consider a neural network designed for $m$-class classification, represented as a series of function compositions:
\begin{equation*}
    f(x) = (\phi^{[l]} \circ \phi^{[l-1]} \circ \ldots \circ \phi^{[1]})(x),
\end{equation*}
where each $\phi^{[i]}$ represents an operation within the network, which could be a linear transformation (such as a fully connected layer), an activation function, or a pooling operation. The parameters of the neural network are collectively denoted as $\theta$. The intermediate outputs of these operations are often referred to as hidden representations. We denote them by $z^{[i]}$, where $z^{[i]} = \phi^{[i]}(z^{[i-1]})$ for $i = 1, 2, \ldots, l$, and the initial input is $z^{[0]} = x$. Additionally, we incorporate the softmax function into the neural network's definition. Specifically, in the final layer, we have $\phi^{[l]} = \text{softmax}(z^{[l-1]})$, where $z^{[l-1]}$ are called the logits. With this definition, the output of this network, $f(x)$, can be interpreted as a probability distribution over the $m$ classes where each component $f(x)_i$ represents the probability of the input $x$ belonging to class $i$.

\subsection{Model Alignment}\label{sec:alignment_formulation}
The goal of model alignment is to modify the source model such that it can extract features similar to those of other models (the witness model). We denote the parameters of the source model and the witness model as $\theta_s$ and $\theta_w$, respectively. Let us consider the following point-wise formulation of the alignment loss:
\begin{equation}\label{eq:alignment_loss_general}
    \ell_a(x, \theta_s, \theta_w) = d(z_s^{[q]}(x), z_w^{[q]}(x)),
\end{equation}
where the metric $d$ measures the output difference at layer $q$ between the models.

Model alignment is a fine-tuning process. During alignment, the parameters of the source model are updated to minimize this alignment loss which captures the differences between models' output. Specifically, when $q=l$, the alignment loss measures the divergence between the probability distributions generated by the two models. In this scenario, the Kullback-Leibler (KL) divergence is particularly suitable, due to its effectiveness in measuring distribution differences and its relative ease of implementation in practice. Our analysis and primary experimental results are based on model alignment in the output space, using KL divergence as $d$. We also explore total variation and alignment in the embedding space (i.e., $q<l$), which is detailed in our ablation study.

The point-wise loss defined in (\ref{eq:alignment_loss_general}) focuses on aligning the source model with a single witness model. However, alignment can also be extended to multiple witness models to further improve the alignment process. This process involves using a set of witness models, denoted as $\Theta$, with the number of witness models represented by $\abs{\Theta}$. Finally, the update rule for the parameters of the source model based on SGD can be written as
\begin{equation*}
    \theta_s(t+1) = \theta_s(t) -  \eta \; \frac{1}{\abs{\mathcal{B}}\abs{\Theta}} \sum_{x\in \mathcal{B}}\sum_{\theta_w\in \Theta} \nabla\ell_a(x, \theta_s(t), \theta_{w}),
\end{equation*}
where $\mathcal{B}$ represents the mini-batch.

\subsection{Understanding Model Alignment}\label{sec:understanding}
We now understand the model alignment approach from two perspectives: semantic feature exploitation and geometric analysis of the loss surface. The analysis in this section uses ResNet50 and ViT-B/16 as source models, which are aligned with ResNet18 and ViT-T/16 as their respective witness models. We consider $\ell_\infty$-norm adversarial examples generated using 20 iterations of PGD~\cite{madry2017towards} with $\eps = 4/255$ and $\alpha=1/255$. We use $\Delta x_s$ and $\Delta x_a$ to represent the perturbations generated based on the source and aligned models, respectively.

\subsubsection{Aligned Model Exploits More Semantic Features.}
To verify that attacks applied to the aligned model exploit more semantic features, we compare the perturbations generated using the original and aligned source models.
\begin{figure}[t]
    \captionsetup[subfigure]{labelformat=simple, labelsep=period}
    \centering 
    \subfloat[ResNet50.]{\includegraphics[width=0.45\linewidth]{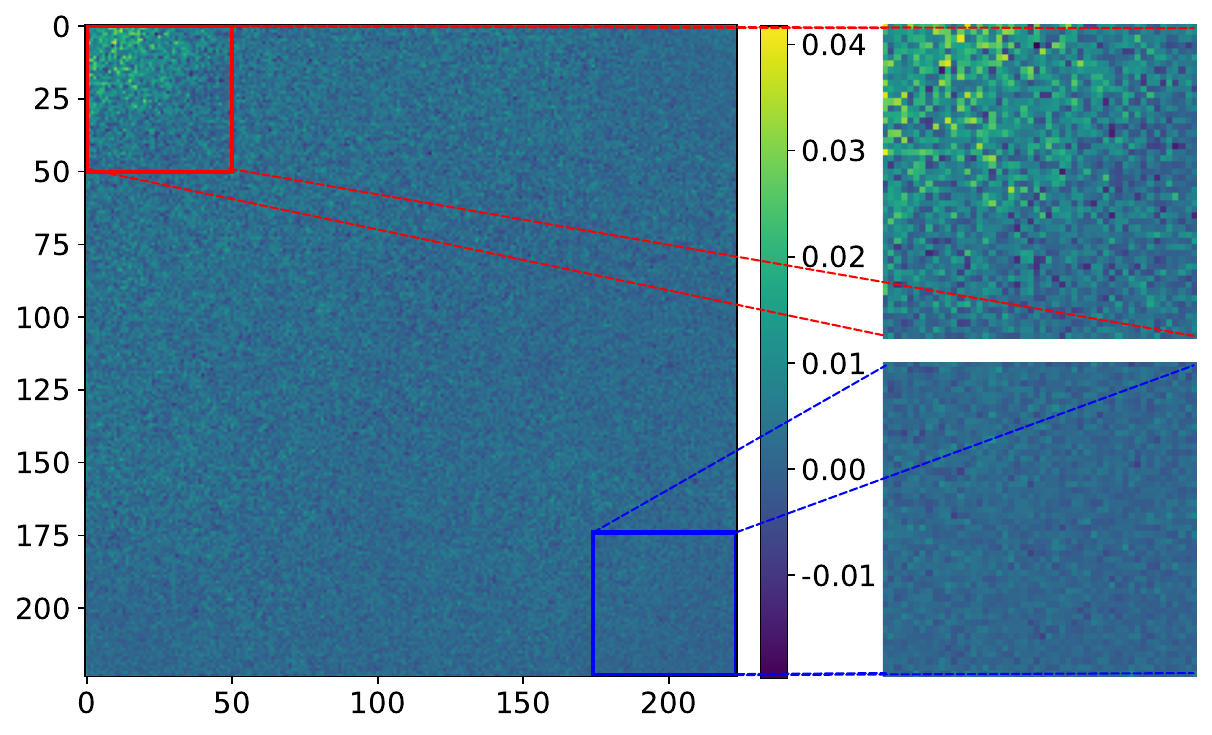}}
    \hspace{0.5cm}
    \subfloat[ViT-B/16.]{\includegraphics[width=0.45\linewidth]{{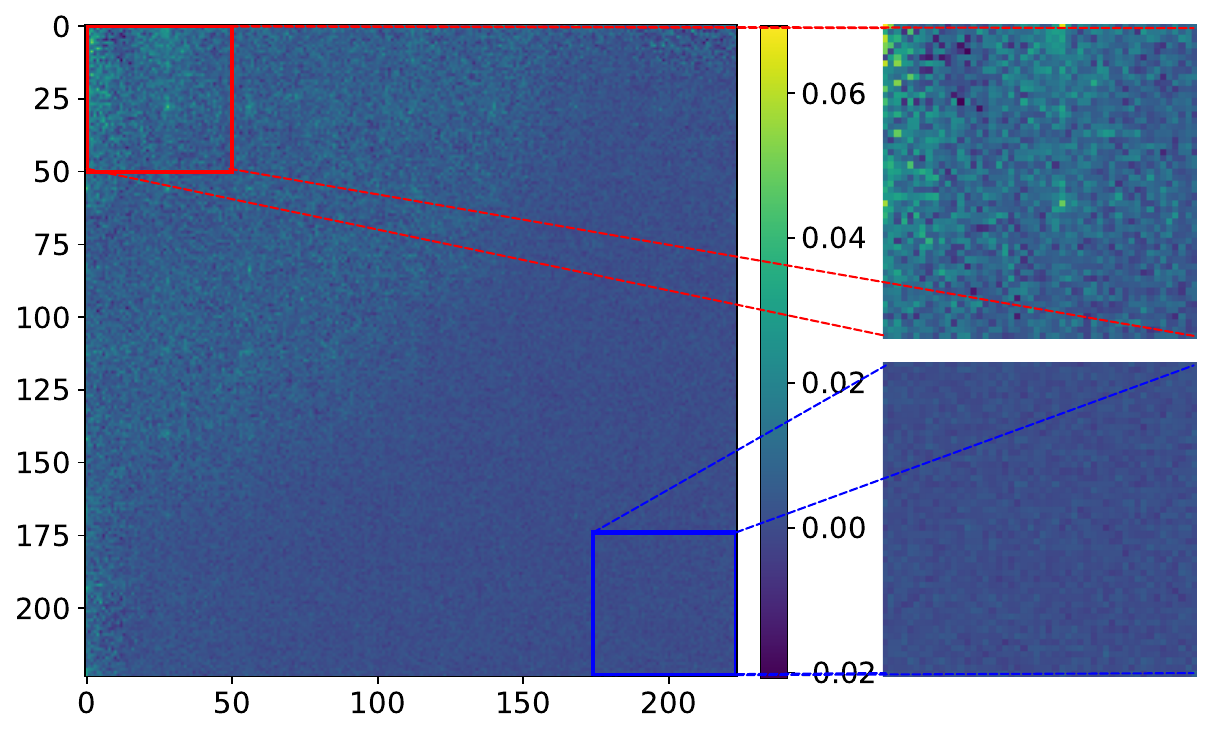}}}
    \caption{\textbf{A frequency-domain visualization of the differences in the perturbation generated using the original source model and the aligned source model.} 
    We compare the magnitude of the DCT coefficients between the perturbations generated by the two models: $\abs{\text{DCT}(\Delta x_a)} - \abs{\text{DCT}(\Delta x_s)}$.
    The pronounced brightness in the top-left region of the spectrum indicates that the primary differences lie within the low-frequency range, which is typically associated with semantic features.} 
    \label{fig:delta_difference_frequency}
\end{figure}
Adversarial perturbations are imperceptible and difficult to characterize in terms of the specific features they exploit in the spatial domain. However, by analyzing perturbations in the frequency domain, we can observe the types of features each model exploits. Previous work has shown that semantic features mostly concentrate around the low-frequency end of the spectrum~\cite{ma2023understanding,wang2020high,yin2019fourier}. We consider discrete cosine transform (DCT)~\cite{ahmed1974discrete} and compare the DCT coefficients of the perturbations generated from the source and the aligned model, denoted as $\text{DCT}(\Delta x_s)$ and $\text{DCT}(\Delta x_a)$, respectively. The results are averaged over 1000 randomly sampled images from the ImageNet test set. Since our interest lies in the magnitude of these coefficients, we visualize the difference by computing $\abs{\text{DCT}(\Delta x_a)} - \abs{\text{DCT}(\Delta x_s)}$.

The results are illustrated in Fig.~\ref{fig:delta_difference_frequency}. We observe that the differences between the perturbations are predominantly located in the top-left corner of the DCT spectrum, indicating that they primarily differ in the amount of low-frequency information. This result shows that adversarial perturbations generated from the aligned exploit more low-frequency, semantic features.

\subsubsection{Model Alignment Yields Smoother Loss Surface.}\label{sec:geometric_analysis}
Previous works have studied the connection between perturbation from sharp loss maxima and their poor transferability~\cite{qin2022boosting, gubri2022lgv}. We extend this geometric perspective by examining how model alignment affects the loss surface geometry of the source model, particularly in its capacity to generate more transferable adversarial examples.

\begin{figure}[t]
    \captionsetup[subfigure]{labelformat=simple, labelsep=period}
    \centering 
    \subfloat[ResNet50.]{\includegraphics[trim={0 0.3cm 22cm 0},clip,width=0.45\linewidth]{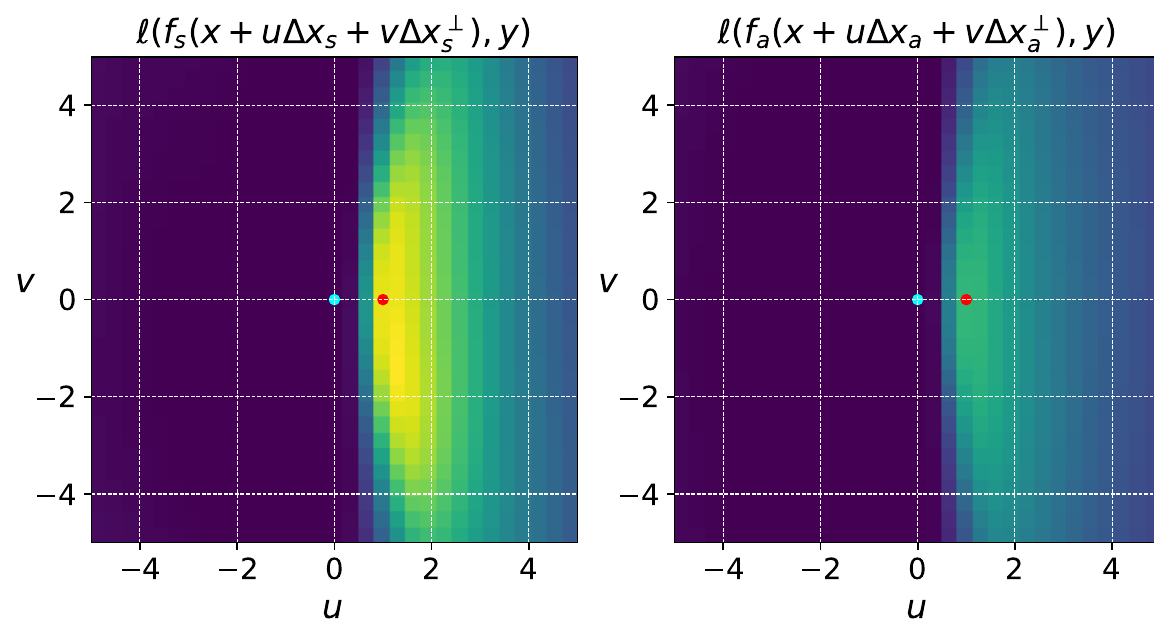}}
    \subfloat[ViT-B/16.]{\includegraphics[trim={0 0.3cm 22cm 0},clip,width=0.46\linewidth]{{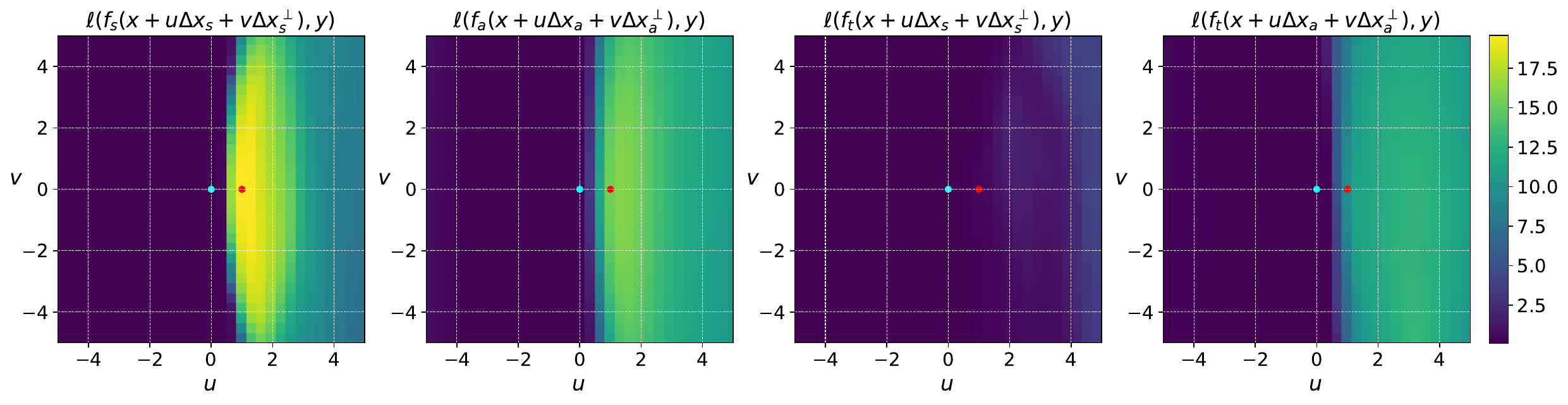}}}
    %
    \caption{\textbf{Visualization of the loss surface around adversarial perturbations for original and aligned ResNet50 and ViT-b/16.} Each plot illustrates the loss surface projected on the plane defined by the adversarial perturbation direction and its orthogonal vector. We examine the loss landscape around a clean data point (cyan) and an $\ell_\infty$-bounded adversarially perturbed data point (red), generated from the source ($\Delta x_s$) and aligned models ($\Delta x_a$). Perturbations from the original source models are at sharper loss maxima, while those from the aligned model are on flatter surfaces.} 
    \label{fig:loss_landscape}
\end{figure}

First, consider when (\ref{eq:alignment_loss_general}) measures the KL divergence between the two models' predictions. During alignment, the source model's parameters are fine-tuned using \textit{soft labels} from the witness model's outputs. Training deep neural networks with soft labels prevents the model from becoming overly confident in its predictions, thereby improving generalization~\cite{szegedy2016rethinking, he2019bag, hinton2015distilling, muller2019does}. Compared to using hard, one-hot encoded labels, it has been shown that training with soft labels implicitly regularizes the norm of the input Jacobian~\cite{carratino2022mixup, zhang2021does}, and leads to smoother decision boundaries~\cite{zhang2018mixup, verma2019manifold}.

To understand why model alignment can be helpful, we begin by identifying a data point for which the perturbation, when generated based on the original source model, fails to transfer to the target model, but the perturbation generated using the aligned model misleads the target model. In Fig.~\ref{fig:loss_landscape}, we visualize the loss surface surrounding this data point, spanned by two pairs of orthogonal vectors:  $\Delta x_s$, $\Delta x_s^{\perp}$, $\Delta x_a$ and $\Delta x_a^{\perp}$. Here, $\Delta x_s^{\perp}$ and $\Delta x_a^{\perp}$ are randomly selected orthogonal vectors and are also bounded by the same $\eps$. In each plot, the center of the plot (cyan) represents the clean data point, and the adversarial example is highlighted using the red marker.

\begin{table}[t]
\begin{center}
\renewcommand{\arraystretch}{1.2}
\renewcommand{\tabcolsep}{6.0pt}
\scriptsize
\caption{ \textbf{Comparing the $\ell_2$-norm of the input gradient on the original source model and the aligned source model.} 
Results are averaged over 1000 randomly selected ImageNet test samples. The norm decreases significantly when evaluated on all three types of input, suggesting that the smoothing effect induced by the alignment is more global, rather than being confined to specific cases of adversarial examples.
}
\label{table:grad_norm_comparison}
\begin{tabular}{ ccccc } 
\Xhline{2\arrayrulewidth}
\multirow{2}{*}{Data}  & \multicolumn{2}{c}{\underline{ResNet50}} & \multicolumn{2}{c}{\underline{VIT-B/16}}\\ 
  & Original & Aligned & Original & Aligned \\\Xhline{2\arrayrulewidth}
Clean  	    & 0.056    & 0.032   & 0.060    & 0.029     \\ 
Gaussian-perturbed	& 0.057    & 0.032   & 0.063    & 0.029     \\
PGD-perturbed			& 0.453    & \gc0.171   & 0.184    & \gc0.055     \\
\Xhline{2\arrayrulewidth}
\end{tabular}
\end{center}
\end{table}
We make two observations. First, the bright yellow region around the perturbations generated from the original source model suggests that they are located at sharp loss maxima. However, these same perturbations do not effectively cause a significant increase in loss when applied to the target model. This observation aligns with previous findings indicating that perturbations with poor transferability often correspond to sharp local maxima unique to the source model, which are not present in the target model~\cite{gubri2022lgv, qin2022boosting}. Second, the loss surface around perturbations generated from the aligned model is noticeably flatter. This is in line with prior findings that adversarial perturbations from flatter maxima tend to be transferable~\cite{gubri2022lgv, qin2022boosting}. It is noteworthy that the PGD attack does not explicitly target flat maxima for generating adversarial examples. This observation leads us to hypothesize that the smoothing effect induced by the alignment process is more global, rather than being confined to specific cases of adversarial examples.

To verify this, we evaluate the change in the $\ell_2$-norm of the loss gradients with respect to clean, Gaussian-perturbed ($\sigma^2 = 0.01$), and PGD-perturbed inputs. More formally, we compare $\norm{\nabla_x \ell(x+\Delta x, y, \theta)}_2$ for $\Delta x \in \set{0, \text{Gaussian}, \text{PGD}}$ for $\theta \in \set{\theta_s, \theta_a}$ over 1000 randomly sampled ImageNet test samples.

The results are summarized in Tab.~\ref{table:grad_norm_comparison}. We observe a significant decrease in gradient norm across all inputs, and the reduction is particularly pronounced at PGD-perturbed data points. This decrease is consistent with the previous finding which demonstrated a connection between the use of soft labels and the smoothing of the loss surface~\cite{carratino2022mixup, zhang2021does}. More importantly, this observation explains the improved transferability with the aligned model. In \ifSupp Appendix~\ref{appendix:input_hessian}\else the supplementary material\fi, we discuss additional results on the change in the largest eigenvalue of input Hessian, which is another popular metric in studying the smoothness of the loss landscape~\cite{zhao2020bridging}. To evaluate the similarity of models, we measure the change in the KL divergence, prediction agreement, and the cosine similarity of input gradients. The latter is crucial due to its role in the attack process. These results are included in \ifSupp Appendix~\ref{appendix:input_hessian}\else the supplementary material\fi.
\section{Experiments}\label{sec:exp}
In this section, we present a series of experiment results to demonstrate the improved ability of a given source model to generate transferable adversarial perturbations. We study factors that can further improve the alignment process and demonstrate the compatibility of our approach with a wide range of attacks.

\subsection{Experiment Setup}
\textbf{Model:} We consider various neural network architectures as source, witness, and target models. For CNNs, we include ResNet18 (Res18), ResNet50 (Res50), ResNet101 (Res101)~\cite{he2016identity}, VGG19~\cite{simonyan2015very}, DenseNet121 (DN121)~\cite{huang2017densely}, and Inception-v3 (IncV3)~\cite{szegedy2016rethinking}. For ViTs, our selection includes Swin Transformers (SWIN)~\cite{liu2021swin}, ViT-T/16, ViT-S/16, and ViT-B/16~\cite{dosovitskiy2020image}. We follow the optimization schedule described in the official Pytorch repository to train all models. Training and model details are included in \ifSupp Appendix~\ref{appendix:training_details}\else the supplementary material\fi.

\noindent\textbf{Fine-tune:} During alignment, all source models are fine-tuned for one epoch using SGD with a momentum of 0.9, a cosine learning rate decay, and a linear warmup. We sweep over 3 learning rates. No additional data is used, as fine-tuning relies on the same training data used to train the source and witness model. We use a batch size of 128 for CNNs and 512 for ViTs. For ViT-based models, we follow~\cite{steiner2021train} and gradients are clipped at a global norm of 1.

\noindent\textbf{Dataset:} We follow previous work \cite{dong2018boosting, lin2019nesterov, yu2023reliable} in which evaluations are based on 1000 randomly selected from the ImageNet test set. For a given source and target model, these samples meet the following criteria: they are correctly classified by both models, and the adversarial examples, when generated from each model, lead to misclassifications in their respective originating models.

\noindent\textbf{Attack method:} We focus on non-targeted adversarial perturbations bounded by the $\ell_\infty$-norm.  Unless otherwise stated, all perturbations are generated using 20 iterations of PGD with $\eps = 4/255$ and $\alpha=1/255$. All target models have an error rate of 100\% under white-box PGD attacks. Results with different attack methods and larger values of $\eps$ are included in the ablation study.

\noindent\textbf{Metric:} We measure transferability using the error rate, with a higher rate indicating greater transferability. In the tables of this section, we first present a row labeled `n/a' to denote the error rate for perturbations generated by the original source model. We then demonstrate the change in transferability after alignment following the $+/-$ sign. A larger change indicates a greater increase in transferability resulting from using the aligned model. All results presented in this section are obtained from the average of three independent runs. 

\subsection{Model Alignment Improves Transferability}\label{sec:exp_diff_arch}
\begin{table}[t]
\begin{center}
\renewcommand{\arraystretch}{1.2}
\renewcommand{\tabcolsep}{2.0pt}
\tiny
\caption{ \textbf{Aligning the source model can lead to the generation of more transferable adversarial examples.} The result demonstrates the increase in error rates when adversarial examples generated from aligned source models are applied to different target models. Our method improves the transferability of adversarial examples generated from a wide range of architectures, including both CNNs and ViTs.
}
\label{table:different_arch}
\begin{tabular}{ cccccccccccc } 
\Xhline{2\arrayrulewidth}
\multirow{2}{*}{Source}		&	\multirow{2}{*}{Witness}	& \multicolumn{10}{c}{Target}\\ \cline{3-12}
&   & Res18 &	Res50 &	Res101 &	VGG19&	DN121&	IncV3&	ViT-T/16&	ViT-S/16&	ViT-B/16&	SWIN \\ \Xhline{2\arrayrulewidth}

\multirow{7}{*}{Res50}
& n/a           & $44.52$	& $61.51$	& $53.01$	& $43.47$	& $50.59$	& $36.05$	& $25.58$	& $21.54$	& $18.86$	& $22.65$ \\
& Res50         & \ul{$+9.60$}	& \gc$+9.90$	& $+8.12$	& $+6.59$	& $+8.98$	& $+4.79$	& $+0.67$	& $+0.65$	& \ul{$+0.42$}	& $+2.65$ \\
& VGG19         & $+8.30$	& $+7.88$	& $+7.41$	& \gc\ul{$+9.17$}	& $+9.93$	& $+6.76$	& $+0.75$	& $+0.52$	& $+0.17$	& $+1.97$ \\
& DN121         & $+9.39$	& \ul{$+10.82$}	& \ul{$+8.40$}	& $+8.41$	& \gc\ul{$+12.41$}	& $+7.21$	& $+1.87$	& \ul{$+2.02$}	& $+0.38$	& $+2.61$ \\
& IncV3         & $+7.69$	& $+7.52$	& $+5.99$	& $+5.56$	& $+9.19$	& \gc\ul{$+8.23$}	& \ul{$+2.29$}	& $+0.30$	& $+0.24$	& \ul{$+3.01$} \\
& ViT-B/16      & $-9.41$	& $-19.39$	& $-17.48$	& $-10.04$	& $-12.76$	& $-8.49$	& $-3.25$	& $-2.69$	& \gc$-1.14$	& $-4.17$ \\
& SWIN          & $-9.23$	& $-16.21$	& $-14.30$	& $-7.27$	& $-10.02$	& $-5.53$	& $-2.72$	& $-2.93$	& $-2.36$	& $-2.73$ \\ \Xhline{2\arrayrulewidth}
\multirow{7}{*}{VGG19}
& n/a           & $33.49$	& $30.10$	& $26.13$	& $79.43$	& $32.07$	& $30.62$	& $23.27$	& $19.85$	& $18.66$	& $23.09$ \\
& Res50         & \ul{$+7.19$}	& \gc\ul{$+4.16$}	& $+2.76$	& $+5.19$	& $+7.62$	& \ul{$+5.91$}	& $+3.45$	& $+1.20$	& $+1.36$	& $+1.12$ \\
& VGG19         & $+4.66$	& $+2.10$	& $+1.77$	& \gc$+7.30$	& $+3.44$	& $+1.74$	& $+2.64$	& $+0.68$	& $+1.39$	& $+0.90$ \\
& DN121         & $+6.35$	& $+3.52$	& \ul{$+5.07$}	& \ul{$+10.60$}	& \gc\ul{$+10.45$}	& $+5.03$	& \ul{$+4.09$}	& \ul{$+2.36$}	& \ul{$+2.58$}	& \ul{$+1.31$} \\
& IncV3         & $+4.91$	& $+1.57$	& $+1.66$	& $+3.82$	& $+4.94$	& \gc$+4.33$	& $+2.70$	& $+0.61$	& $+0.61$	& $+1.04$ \\
& ViT-B/16      & $-4.34$	& $-5.68$	& $-5.01$	& $-24.66$	& $-5.11$	& $-3.67$	& $-0.83$	& $-0.37$	& \gc$+0.77$	& $-2.53$ \\
& SWIN          & $-3.24$	& $-5.83$	& $-4.01$	& $-21.91$	& $-6.48$	& $-4.05$	& $-1.11$	& $-0.73$	& $+0.64$	& $-2.96$ \\ \Xhline{2\arrayrulewidth}
\multirow{7}{*}{DN121}
& n/a           & $40.20$	& $44.01$	& $37.15$	& $40.84$	& $61.15$	& $32.28$	& $23.68$	& $20.34$	& $18.21$	& $23.56$ \\
& Res50         & $+11.67$	& \gc\ul{$+12.96$}	& $+8.37$	& $+7.32$	& $+11.68$	& $+5.41$	& $+1.69$	& $+0.46$	& $+1.00$	& $+1.59$ \\
& VGG19         & $+9.02$	& $+6.38$	& $+4.88$	& \gc\ul{$+12.34$}	& $+7.87$	& $+3.77$	& $+2.72$	& $+0.04$	& $+1.22$	& $+0.94$ \\
& DN121         & \ul{$+12.56$}	& $+11.60$	& \ul{$+10.48$}	& $+9.10$	& \gc\ul{$+15.47$}	& $+6.84$	& \ul{$+3.99$}	& \ul{$+1.59$}	& \ul{$+1.24$}	& $+1.86$ \\
& IncV3         & $+8.08$	& $+10.50$	& $+9.88$	& $+9.91$	& $+12.76$	& \gc\ul{$+7.24$}	& $+3.81$	& $+1.23$	& $-0.02$	& \ul{$+2.79$} \\
& ViT-B/16      & $-3.08$	& $-8.43$	& $-7.78$	& $-6.67$	& $-12.96$	& $-1.78$	& $+2.06$	& $+1.26$	& \gc$+1.22$	& $-2.78$ \\
& SWIN          & $-4.13$	& $-8.57$	& $-5.34$	& $-6.11$	& $-10.21$	& $-4.64$	& $+0.54$	& $-0.16$	& $+1.21$	& $-1.88$ \\ \Xhline{2\arrayrulewidth}
\multirow{7}{*}{IncV3}
& n/a           & $30.68$	& $27.49$	& $25.02$	& $33.47$	& $30.98$	& $51.80$	& $22.99$	& $18.77$	& $17.72$   & $20.94$ \\
& Res50         & \ul{$+4.92$}	& \gc\ul{$+6.59$}	& \ul{$+4.27$}	& $+5.49$	& $+5.74$	& $+4.49$	& $+1.71$	& \ul{$+2.26$}	& \ul{$+2.74$}	& \ul{$+2.23$} \\
& VGG19         & $+4.65$	& $+2.69$	& $+3.10$	& \gc$+5.70$	& $+3.39$	& $+4.36$	& $+1.05$	& $-0.54$	& $+0.38$	& $+0.53$ \\
& DN121         & $+4.50$	& $+6.38$	& $+2.13$	& \ul{$+7.35$}	& \gc\ul{$+6.52$}	& \ul{$+8.89$}	& \ul{$+3.54$}	& $+0.68$	& $+1.41$	& $+1.45$ \\
& IncV3         & $+1.78$	& $+3.91$	& $+3.70$	& $+3.53$	& $+3.06$	& \gc$+8.75$	& $+1.79$	& $+0.49$	& $+1.41$	& $+1.75$ \\
& ViT-B/16      & $+0.47$	& $-1.13$	& $-2.97$	& $-0.77$	& $-1.35$	& $-8.20$	& $+2.92$	& $+1.64$	& \gc$+2.07$	& $-0.23$ \\
& SWIN          & $-0.49$	& $-3.36$	& $-3.85$	& $-3.29$	& $-4.54$	& $-12.82$	& $+0.09$	& $+0.76$	& $+1.13$	& $-1.34$ \\ \Xhline{2\arrayrulewidth}
\multirow{7}{*}{ViT-B/16}
& n/a           & $23.05$	& $19.67$	& $18.30$	& $21.83$	& $21.96$	& $ 21.52$	& $36.44$	& $44.58$	& $51.27$	& $21.22$ \\
& Res50         & \ul{$+17.53$}	& \gc$+12.89$	& $+12.23$	& $+14.83$	& $+16.66$	& $+15.76$	& $+43.47$	& $+37.17$	& $+29.16$	& $+21.16$ \\
& VGG19         & $+10.89$	& $+9.49$	& $+9.77$	& \gc$+14.23$	& $+12.65$	& $+10.94$	& $+39.63$	& $+31.65$	& $+21.30$	& $+14.97$ \\
& DN121         & $+17.15$	& \ul{$+13.92$}	& \ul{$+12.28$}	& \ul{$+15.38$}	& \gc\ul{$+17.68$}	& \ul{$+16.96$}	& \ul{$+47.36$}	& \ul{$+41.88$}	& \ul{$+31.22$}	& \ul{$+21.87$} \\
& IncV3         & $+15.41$	& $+11.87$	& $+11.31$	& $+15.09$	& $+14.23$	& \gc$+16.46$	& $+46.07$	& $+40.39$	& $+28.87$	& $+19.92$ \\
& ViT-B/16      & $+5.14$	& $+2.04$	& $+2.07$	& $+4.79$	& $+2.87$	& $+4.53$	& $+24.04$	& $+23.14$	& \gc$+21.93$	& $+6.34$ \\
& SWIN          & $+6.31$	& $+5.75$	& $+5.92$	& $+8.12$	& $+7.46$	& $+7.28$	& $+39.11$	& $+36.49$	& $+28.61$	& $+14.62$ \\ \Xhline{2\arrayrulewidth}
\end{tabular}
\end{center}
\end{table}
To demonstrate the improved transferability of perturbations generated using the aligned source model, we evaluate across a diverse set of neural network architectures for both source and witness models, focusing on the transferability to an extensive array of target models. Tab.~\ref{table:different_arch} shows the original error rates and their changes after using the aligned model. We make two key observations.

First, the model alignment approach can transform any given source model into one from which the PGD attack generates more transferable perturbations. Such an improvement can be observed across various combinations of source and witness models and is evident in all evaluated target models. This clear improvement validates our approach, showing that aligning the source model can lead to the generation of more transferable adversarial perturbations.
Notably, both CNN and ViT-based source models can benefit from the alignment process.
Recent studies have shown that the transferability between ViTs and CNNs is poor, and many attack algorithms do not generalize well to ViTs~\cite{naseer2021improving,wei2022towards,mahmood2021robustness}.
Our model alignment method offers a promising solution to bridge this gap.

Second, we find that CNN-based source models generally benefit from aligning with other CNN-based models, rather than with ViT-based models.
On the other hand, the ViT-B/16 source model can benefit from aligning with witness models from both CNN and ViT families.
We provide a possible explanation for this phenomenon.
Previous studies have shown that ViTs and CNNs learn distinctively different features~\cite{zhou2022understanding, raghu2021vision}.
Specifically, Raghu \etal~\cite{raghu2021vision} focused on early-layer representations learned by the two models and showed that more ResNet layers are required to be modified to match hidden representations of a ViT, compared to the other way around.
In the context of model alignment, this suggests an asymmetric behavior in the alignment process; namely, it might be easier to align ViTs with CNNs than to align CNNs with ViTs.

\textbf{Choosing the witness model:}
The main goal of Tab.~\ref{table:different_arch} is to demonstrate that the improved transferability is not limited to particular choices of witness models. The result also highlights the effectiveness of a straightforward, hyper-parameter-free \textbf{self-alignment} strategy. This is evident from the consistent improvement observed when the source and witness models share the same architecture, but are initialized and trained independently. The selection of witness models is further explored in our ablation studies.

\textbf{Regularization to prevent overfitting:}
A prolonged alignment process may inadvertently cause overfitting to the witness model, thereby diminishing the gains in transferability. To counteract overfitting, regularization methods can be implemented. For example, we apply early stopping and limit the alignment to a single epoch. Moreover, an ensemble of witnesses can be used to avert overfitting to any singular witness model. Results with multiple witness models are included in \ifSupp Appendix~\ref{appendix:exp_num_witness}\else the supplementary material\fi. Furthermore, when aligning by minimizing KL divergence, adjusting the temperature scaling within the softmax function can be an effective measure to prevent the exact replication of the witness model's predictions by the source model.

\subsection{Ablation Studies}\label{sec:exp_ablation}
We conduct several ablation studies to investigate factors that could further improve the alignment process. Additionally, we demonstrate that our approach is compatible with a wide range of attack algorithms.

\subsubsection{Smaller Witness Model Might Boost Learning Shared Features.}
Results in Tab.~\ref{table:different_arch} indicate that DN121, which has the fewest parameters, frequently emerges as a more effective witness model. This motivates us to investigate the role of the capacity of the witness model during the alignment. While several factors contribute to a model's capacity, such as its structure, the normalization techniques, and its non-linear activations, our study primarily focuses on the number of parameters as a proxy for capacity. With this in mind, we consider three models each from the ResNet and ViT families.

Tab.~\ref{table:model_capacity} summarizes the results of the model capacity analysis, pointing to one key observation. Model alignment is more effective when the witness models have a smaller model capacity. For instance, when using Res101 as the source model, alignment with the lower capacity Res18 as the witness model is more beneficial than alignment with Res50. This matches the previous findings in Tab.~\ref{table:different_arch}.

We provide one interpretation of this observation. Smaller models might tend to focus more on learning semantic features for generalization, as they lack the capacity of larger models to learn imperceptible features. Therefore, when a source model is aligned with a smaller model, it is steered towards learning more semantic features that are commonly shared across different models, thereby leading to more transferable perturbations.

\begin{table*}[t]
\begin{center}
\renewcommand{\arraystretch}{1.2}
\renewcommand{\tabcolsep}{2.0pt}
\tiny
\caption{ \textbf{Analyzing the impact of witness model capacity on the alignment process. } We observe a greater improvement in the transferability of adversarial perturbations when the source model is aligned with witness models of smaller capacity.}
\label{table:model_capacity}
\begin{tabular}{ cccccccccccc }
\Xhline{2\arrayrulewidth}
\multirow{2}{*}{Source}		&	\multirow{2}{*}{Witness}	& \multicolumn{10}{c}{Target}\\ \cline{3-12}
  &   & Res18 &	Res50 &	Res101 &	VGG19&	DN121&	IncV3&	ViT-T/16&	ViT-S/16&	ViT-B/16&	SWIN \\ \Xhline{2\arrayrulewidth}

\multirow{3}{*}{Res18}
& n/a       & $64.10$	& $45.85$	& $37.09$	& $42.96$	& $48.57$	& $37.31$	& $27.52$	& $21.12$	& $20.57$	& $21.31$ \\
& Res50     & $-1.94$	& $+0.41$	& $-0.74$	& $+0.82$	& $-1.24$	& $-1.78$	& $+0.26$	& $+1.33$	& $+0.42$	& $+1.48$ \\
& Res101    & $-5.13$	& $-3.92$	& $-2.70$	& $-1.75$	& $-1.39$	& $-0.22$	& $+0.53$	& $+0.25$	& $+0.03$	& $+0.48$ \\
\hline
\multirow{3}{*}{Res50}
& n/a       & $44.52$	& $61.51$	& $53.01$	& $43.47$	& $50.59$	& $36.05$	& $25.58$	& $21.54$	& $18.86$	& $22.65$ \\
& Res18     & \gc$+35.30$	& \gc$+26.56$	& \gc$+25.20$	& \gc$+24.38$	& \gc$+27.86$	& \gc$+19.18$	& \gc$+5.49$	& \gc$+2.85$	& \gc$+1.61$	& \gc$+9.60$ \\
& Res101    & $+4.27$	& $+3.02$	& $+3.93$	& $+3.61$	& $+4.41$	& $+3.60$	& $+1.86$	& $-0.54$	& $+0.14$	& $+1.36$ \\
\hline
\multirow{3}{*}{Res101}
& n/a       & $38.84$	& $54.63$	& $62.95$	& $39.24$	& $49.93$	& $34.77$	& $23.01$	& $20.12$	& $19.07$	& $22.90$ \\
& Res18     & \gc$+40.64$	& \gc$+35.11$	& \gc$+27.78$	& \gc$+29.63$	& \gc$+33.20$	& \gc$+28.42$	& \gc$+10.10$	& \gc$+5.70$	& \gc$+2.37$	& \gc$+13.99$ \\
& Res50     & $+14.97$	& $+19.56$	& $+16.97$	& $+13.50$	& $+17.07$	& $+10.41$	& $+4.04$	& $+0.98$	& $+1.44$	& $+4.81$ \\
\Xhline{2\arrayrulewidth}
\multirow{3}{*}{ViT-T/16}
& n/a       & $28.77$	& $23.45$	& $21.05$	& $25.41$	& $25.45$	& $25.22$	& $63.89$	& $40.47$	& $29.62$	& $22.97$ \\
& ViT-S/16  & $+0.30$	& $+0.70$	& $+1.02$	& $+2.76$	& $+0.54$	& $+0.69$	& $+7.01$	& $+3.77$	& $+2.92$	& $+0.42$ \\
& ViT-B/16  & $+0.86$	& $-0.18$	& $-1.11$	& $+1.15$	& $+0.47$	& $+0.81$	& $+5.09$	& $+4.32$	& $+3.55$	& $-1.57$ \\
\hline
\multirow{3}{*}{ViT-S/16}
& n/a       & $24.27$ 	& $20.48$	& $18.86$	& $21.85$	& $21.90$	& $22.96$	& $47.34$	& $52.09$	& $43.85$	& $23.87$ \\
& ViT-T/16  & \gc$+9.64$	& \gc$+6.65$	& \gc$+6.29$	& \gc$+8.51$	& \gc$+10.66$	& \gc$+7.58$	& \gc$+40.53$	& \gc$+32.27$	& \gc$+26.10$	& \gc$+12.35$ \\
& ViT-B/16  & $+3.10$	& $+1.21$	& $+2.87$	& $+3.75$	& $+3.19$	& $+3.36$	& $+19.07$	& $+17.44$	& $+13.81$	& $+4.73$ \\
\hline
\multirow{3}{*}{ViT-B/16}
& n/a       & $23.05$	& $19.67$	& $18.30$	& $21.83$	& $21.96$	& $21.52$	& $36.44$	& $44.58$	& $51.27$	& $21.22$ \\
& ViT-T/16  & \gc$+10.13$	& \gc$+9.71$	& \gc$+8.42$	& \gc$+10.71$	& \gc$+11.30$	& \gc$+9.89$	& \gc$+51.72$	& \gc$+44.99$	& \gc$+34.69$	& \gc$+20.01$ \\
& ViT-S/16  & $+7.30$	& $+4.57$	& $+4.53$	& $+4.75$	& $+6.21$	& $+6.34$	& $+36.47$	& $+35.66$	& $+29.43$	& $+12.00$ \\
\Xhline{2\arrayrulewidth}
\end{tabular}
\end{center}
\end{table*}
\begin{table*}[t]
\begin{center}
\renewcommand{\arraystretch}{1.2}
\renewcommand{\tabcolsep}{3.5pt}
\tiny
\caption{ \textbf{Improving transferability via embedding-space alignment.} We focus on Res50 as the source model and Res18 as the witness model. We observe improved transferability when alignment is performed in the embedding space or when combining output and embedding spaces.}
\label{table:kd_method_res}
\begin{tabular}{ ccccccccccc } 
\Xhline{2\arrayrulewidth}
\multirow{2}{*}{Method}	& \multicolumn{10}{c}{Target}\\ \cline{2-11}
& Res18 &	Res50 &	Res101 &	VGG19&	DN121&	IncV3&	ViT-T/16&	ViT-S/16&	ViT-B/16&	SWIN \\ \Xhline{2\arrayrulewidth}

n/a 							& $44.52$	      & $61.51$	    & $53.01$	     & $43.47$	     & $50.59$	     & $36.05$	    & $25.58$	    & $21.54$	    & $18.86$	    & $22.65$ \\
KL 								& $+35.30$	      & \gc$+26.56$	& $+25.20$	     & $+24.38$	     & $+27.86$	     & $+19.18$	    & $+5.49$	    & $+2.85$	    & $+1.61$	    & $+9.60$ \\
TV 								& $+26.04$	      & $+19.91$	& $+11.88$	     & $+17.40$	     & $+20.66$	     & $+15.93$	    & $+5.70$	    & $+1.72$	    & $+0.85$	    & $+5.79$ \\\hline
RKD~\cite{park2019relational} 	& $+2.95$	      & $+2.12$	    & $+2.80$	     & $+2.78$	     & $+3.35$	     & $+4.09$	    & $-1.26$	    & $-0.53$	    & $-0.36$	    & $+0.21$ \\
HINT~\cite{romero2014fitnets} 	& $+5.95$	      & $+6.55$	    & $+5.27$	     & $+5.43$	     & $+9.54$	     & $+4.34$	    & $+2.55$	    & $-0.01$	    & $+1.24$	    & $+2.58$ \\
EGA~\cite{ma2022distilling} 	& $+35.35$	      & $+23.56$	& \gc$+27.09$	 & $+24.69$	     & $+28.94$	     & \gc$+22.00$	& \gc$+10.12$	& \gc$+4.62$	& \gc$+3.80$	& $+10.37$ \\
NCE~\cite{chen2021distilling} 	& $+21.51$	      & $+19.25$	& $+20.39$	     & $+16.06$	     & $+19.70$	     & $+13.29$	    & $+6.78$	    & $+3.38$	    & $+1.06$	    & $+6.73$ \\\hline
KL+EGA 							& \gc$+35.74$	  & $+26.10$	& $+23.23$	     & \gc$+28.90$	 & \gc$+29.59$	 & $+20.06$	    & $+8.62$	    & $+3.81$	    & $+1.81$	    & \gc$+14.05$ \\
KL+NCE                          & $+31.35$	      & $+24.46$	& $+20.12$	     & $+22.62$	     & $+27.14$	     & $+19.06$	    & $+8.48$	    & $+1.50$	    & $+2.36$	    & $+12.18$ \\
\Xhline{2\arrayrulewidth}
\end{tabular}
\end{center}
\end{table*}
\subsubsection{Alignment in the Embedding Space Can Further Improve Transferability.}\label{sec:exp_kl_method}
Recent advances in knowledge distillation highlight the benefits of aligning intermediate representations over aligning outputs~\cite{ma2022distilling,chen2021distilling,park2019relational}. Motivated by this, we investigate aligning the hidden representations of a Res50 source model with a Res18 witness model. We set $q$ in (\ref{eq:alignment_loss_general}) to the layer before the fully-connected layer and evaluate four embedding-space distillation methods: Relational Knowledge Distillation (RKD)~\cite{park2019relational}, Embedding Graph Alignment (EGA)~\cite{ma2022distilling}, Intermediate-level Hints (HINT)~\cite{romero2014fitnets} and Noise Contrastive Estimation (NCE)~\cite{chen2021distilling}. These methods correspond to different choices of distance metric $d$ in (\ref{eq:alignment_loss_general}). Moreover, alignment can be performed in both output and embedding spaces, requiring the optimization of a combined objective with a scaling hyper-parameter ($\lambda$): $d_q(z_s^{[q]}(x), z_w^{[q]}(x)) + \lambda d_p(z_s^{[p]}(x), z_w^{[p]}(x))$, where $p<q=l$. In our experiment, we combine KL with EGA and NCE ($\lambda = 1$). 

The results are summarized in Tab.~\ref{table:kd_method_res}, where we also supplement the previous KL-based results with total variation (TV). These results demonstrate improved transferability with other alignment formulations, including both embedding-space and output-embedding-space alignment. These promising findings pave the way for further exploration of embedding-space model alignment to improve adversarial transferability..

\begin{table}[t]
\begin{center}
\renewcommand{\arraystretch}{1.2}
\renewcommand{\tabcolsep}{2.7pt}
\tiny
\caption{ \textbf{Compatibility of model alignment with different attack algorithms. } 
The Res50 and ViT-B/16 source models are aligned with a Res18 and ViT-T/16 witness model, respectively. Each entry begins with the error rate for perturbations generated by the original source model. Then, this is followed by the change in the error rate after model alignment. The results demonstrate that the model alignment process enhances the effectiveness of all attack algorithms.}
\label{table:orthogonality}
\begin{tabular}{ cccccccc } 
\Xhline{2\arrayrulewidth}
\multirow{2}{*}{Source}		&	\multirow{2}{*}{Attack}	& \multicolumn{6}{c}{Target}\\ \cline{3-8}
  &   & Res18 & VGG19&	DN121&	IncV3&		ViT-B/16&	SWIN \\ \Xhline{2\arrayrulewidth}

\multirow{7}{*}{Res50}
& MI~\cite{dong2018boosting} 	& $75.54+20.23$ & $71.46+19.97$ & $79.60+16.31$ & $59.08+22.22$& $25.19+7.87$ & $37.04+16.02$ \\ 
& NI~\cite{lin2019nesterov} 	& $59.71+21.99$ & $56.30+19.14$ & $59.56+19.90$ & $46.45+15.59$& $26.66+3.62$ & $28.62+8.41$ \\ 
& VMI~\cite{wang2021enhancing} 	& $85.26+12.70$& $83.16+13.13$ & $88.09+9.72$ & $71.98+16.68$ & $29.62+8.32$ & $49.13+15.45$ \\ 
& VNI~\cite{wang2021enhancing} 	& $86.92+11.18$& $85.64+11.36$ & $90.47+8.21$ & $73.60+16.40$ & $29.15+8.58$ & $46.64+16.75$ \\ 
& SINI~\cite{lin2019nesterov} 	& $67.43+17.05$& $61.73+11.87$ & $65.82+13.34$ & $53.74+11.96$ & $26.07+3.03$ & $30.59+5.29$ \\ 
& TI~\cite{dong2019evading} 	& $71.31+21.67$ & $64.87+20.13$ & $73.95+16.85$ & $53.13+19.18$ & $25.58+7.59$ & $30.00+11.71$ \\ 
& DI~\cite{xie2019improving} 	& $87.95+11.03$& $82.12+15.88$ & $90.02+9.40$ & $77.78+15.45$ & $27.48+8.69$ & $43.20+21.83$ \\ 

\Xhline{2\arrayrulewidth}
\multirow{7}{*}{ViT-B/16}
& MI~\cite{dong2018boosting} 	& $33.73+15.60$ & $35.24+12.06$ & $32.29+13.51$ & $33.97+11.85$ & $71.48+24.13$ & $33.57+20.87$ \\ 
& NI~\cite{lin2019nesterov} 	& $35.10+6.93$ & $35.58+4.58$ & $31.93+7.44$ & $33.67+5.70$ & $53.55+17.73$ & $30.69+8.60$ \\ 
& VMI~\cite{wang2021enhancing} 	& $35.59+18.65$ & $37.82+14.72$ & $34.55+16.12$ & $34.42+15.15$ & $77.20+19.14$ & $37.63+21.75$ \\ 
& VNI~\cite{wang2021enhancing} 	& $33.99+22.45$ & $37.19+17.16$ & $32.59+19.92$ & $33.19+20.83$ & $80.07+17.44$ & $38.19+21.89$ \\ 
& SINI~\cite{lin2019nesterov} 	& $36.68+11.40$& $36.56+7.49$ & $33.12+9.63$ & $35.87+9.70$ & $62.07+9.97$ & $33.02+10.26$ \\ 
& TI~\cite{dong2019evading} 	& $36.85+29.96$ & $32.74+24.05$ & $35.48+25.98$ & $32.99+21.31$ & $67.16+27.23$ & $32.50+25.52$ \\ 
& DI~\cite{xie2019improving} 	& $41.75+30.79$ & $39.03+30.42$ & $41.40+26.69$ & $38.62+30.51$ & $84.10+13.70$ & $42.09+29.79$ \\ 
\Xhline{2\arrayrulewidth}
\end{tabular}
\end{center}
\end{table}
\begin{table}[t]
\begin{center}
\renewcommand{\arraystretch}{1.2}
\renewcommand{\tabcolsep}{2.4pt}
\tiny
\caption{\textbf{Improved transferability from using aligned models alone or in an ensemble.} Perturbations from the aligned model (Res50*) demonstrate higher transferability compared to those from an ensemble of source and witness models.}
\label{table:ensemble}
\begin{tabular}{ cccccccccccc }
\Xhline{2\arrayrulewidth}
\multirow{2}{*}{Source}  & \multicolumn{10}{c}{Target}\\ \cline{2-12}
& Res18 & Res50 &  Res101 & VGG19& DN121& IncV3& ViT-T/16& ViT-S/16& ViT-B/16& SWIN \\ \Xhline{2\arrayrulewidth}

Res18  			& 87.53 & 67.411 & 52.074 & 58.85 & 67.12 & 50.22 & 31.88 & 24.65 & 19.16 & 30.91  \\
Res50  			& 66.97 & 85.02 & 72.14 & 64.88 & 75.89 & 49.56 & 32.14 & 25.23 & 22.58 & 28.96  \\ \hline
Res50* 			& \gc{92.03} & \gc{95.73} & \gc{89.86} & \gc{90.60} & \gc{92.51} & \gc{71.42} & \gc{44.12} & \gc{33.24} & \gc{25.51} & \gc{49.12}  \\
Res50 + Res18~\cite{dong2018boosting}   & 91.25 & 91.56 & 80.60 & 75.90 & 85.36 & 60.11 & 19.01 & 15.95 & 10.90 & 27.10  \\ \hline
2$\times$Res50*~\cite{dong2018boosting} & \gc{98.70} & \gc{99.44} & \gc{98.32} & \gc{96.09} & \gc{98.71} & \gc{84.65} & \gc{38.98} & \gc{25.71} & \gc{15.42} & \gc{57.30}  \\ 
2$\times$(Res50 + Res18)~\cite{dong2018boosting}  & 97.69 & 98.73 & 96.81 & 92.95 & 98.32 & 78.61 & 23.87 & 18.06 & 12.25 & 41.77  \\
\Xhline{2\arrayrulewidth}
\end{tabular}
\end{center}
\end{table}
\subsubsection{Model Alignment is Compatible with Other Transfer-enhancing Methods.}
To demonstrate that model alignment is compatible with a wide range of attack algorithms, we extend our analysis to include additional transfer-enhancing attacks. 
This includes optimization-based methods such as MI-FGSM~\cite{dong2018boosting}, NI-FGSM/SINI-FGSM~\cite{lin2019nesterov} and VMI-FGSM/VNI-FGSM~\cite{wang2021enhancing}. We also include data-augmentation-based methods such as TI-FGSM~\cite{dong2019evading} and DI-FGSM~\cite{xie2019improving}. Moreover, we consider a larger $\ell_\infty$-norm constraint of $\eps=8/255$ with a step size of $\alpha=2/255$, and double the number of iterations to 40. The results are summarized in Tab.~\ref{table:orthogonality}. Each entry begins with the error rate for perturbations generated by the original source model. Then, this is followed by the change in the error rate after model alignment. We observe that all considered attack algorithms can leverage the aligned source model to improve transferability. 

\textbf{Model-modification attacks:} In addition to the attacks included in the table, we show that model alignment can be integrated with LinBP~\cite{guo2020backpropagating} and BPA~\cite{xiaosen2024rethinking}, two model-modification-based methods. We compare the transferabilty of MI-FGSM-LinBP and PGD-BPA perturbations from the aligned Res50 to those from the original, unaligned Res50. We observe increased transferability across all target models, with an average increase of $9.66\%$ and $16.86\%$, respectively. These results are included in \ifSupp Appendix~\ref{appendix:model_modification}\else the supplementary material\fi.

\textbf{Ensemble attacks:} Given that an additional witness model is involved in the attack process, it is important to compare perturbations produced by an aligned model with those generated by an ensemble of the original source and witness model. In Tab.~\ref{table:ensemble}, we first demonstrate transferability using the original Res18 and Res50, followed by a Res50 aligned with Res18~(Res50*). Next, we incorporate the ensemble in logits scheme~\cite{dong2018boosting} into the PGD attack, and consider three scenarios: an ensemble of the source and witness models~(Res50+Res18), an ensemble of two aligned models~(2$\times$Res50*), and an ensemble of two pairs of source-witness models~(2$\times$(Res50+Res18)).

We make two observations. First, perturbations from a single aligned model demonstrate higher transferability compared to those from an ensemble of source and witness models, and an ensemble of two aligned models not only further improves transferability but also surpasses the performance of the four-model ensemble. Beyond the improved transferability, model alignment brings the added benefits of faster inference times and lower memory requirements, presenting a significant advantage over conventional ensemble approaches. Second, when targeting ViTs, perturbations generated from ResNet ensembles exhibit limited transferability. This observation is in line with the previous result~\cite{ma2023transferable}, underscoring the specific challenges in transferring attacks between different architectures.
\section{Conclusions}
In this paper, we proposed model alignment as a novel perspective in improving the transferability of adversarial examples.
During alignment, the parameters of the source model are fine-tuned to minimize an alignment loss which measures the divergence in the predictions between the source and the witness model.
We conduct a geometric analysis to study the changes in the loss landscape resulting from this process to better understand the underlying effect of model alignment.
Extensive experiments on the ImageNet dataset demonstrate that perturbations generated from aligned source models exhibit significantly higher transferability than those from the original source model.
A limitation of our study is the absence of a theoretical framework to understand the model alignment process, and we consider it as future work.
%


\subsubsection*{Acknowledgments.}
Avery Ma acknowledges the funding from the Natural Sciences and Engineering Research Council (NSERC) through the Canada Graduate Scholarships – Doctoral (CGS D) program. 
Amir-massoud Farahmand acknowledges the funding from the CIFAR AI Chairs program, as well as the support of the NSERC through the Discovery Grant program (2021-03701). 
Yangchen Pan, Philip Torr and Jindong Gu acknowledge the support from the UKRI Grant: Turing AI Fellowship
EP/W002981/1, EPSRC/MURI Grant: EP/N019474/, and the Royal Academy of Engineering. 
Resources used in preparing this research were provided, in part, by the Province of Ontario, the Government of Canada through CIFAR, and companies sponsoring the \href{http://www.vectorinstitute.ai/partners}{Vector Institute}. 
We would like to also thank the members of the Adaptive Agents Lab who provided feedback on a draft of this paper.

%
%
\bibliographystyle{splncs04}
\bibliography{main}

\appendix
\ifSupp 
\clearpage
\section{Summary of the Supplementary Material}
The supplementary material is organized as follows. In Appendix~\ref{appendix:training_details}, we first describe the implementation details, including the exact optimization schedule, model architectures, data augmentations, attack algorithms, and methods for alignment in the embedding space. In Appendix~\ref{appendix:understanding}, we discuss additional empirical studies on understanding the model alignment process. Finally, additional experiment results and discussions are included in Appendix~\ref{appendix:exp}.

\section{Implementation Details}\label{appendix:training_details}
For all models considered in our paper, we adhere to the optimization configurations as detailed in the official PyTorch repository\footnote{\url{https://github.com/pytorch/vision/tree/main/references/classification}}.

\noindent\textbf{CNN-based models:} All CNN-based models are trained using the same configurations. Those models include ResNet18 (Res18), ResNet50 (Res50), ResNet101 (Res101), VGG19, DenseNet121 (DN121), and Inception-v3 (IncV3). They are trained for 90 epochs using SGD with an initial learning rate of 0.1 and a momentum coefficient of 0.9. We start training with a 5-epoch learning rate warmup, followed by a cosine decay schedule. The batch size is set at 256. 

\noindent\textbf{ViT-based models:} The ViT-based models include ViT-T/16, ViT-S/16, ViT-B/16, and Swin Transformers (SWIN). They are all trained for 300 epochs using AdamW. The initial learning rates are set at 0.003 for ViT models and 0.001 for SWIN. For the ViT's, training begins with a 30-epoch learning rate warmup, followed by a cosine decay schedule, whereas for SWIN, the warmup is 20 epochs. The batch size is set at 1024. We use label smoothing during training. To ensure training stability, the global gradient norm is clipped at 1.

\noindent\textbf{Data augmentations:} For the training of all CNN-based models, augmentation techniques are random resizing, cropping, and flipping. For the training of all ViT-based models, we further incorporate Mixup~\cite{zhang2018mixup} and Cutmix~\cite{yun2019cutmix}. Only random resizing, cropping and flipping are used during the alignment process. Specifically, RandAugment~\cite{cubuk2020randaugment} is applied to ViT's, while TrivialAugment~\cite{muller2021trivialaugment} and random erasing~\cite{zhong2020random} are applied to SWIN. 

\noindent\textbf{Model definitions:} All model definitions are obtained from the torchvision library~\cite{torchvision2016}, with the exceptions of IncV3, ViT's, and SWIN, which are obtained from the timm library~\cite{rw2019timm}.

\noindent\textbf{Attack algorithms:} All the attack algorithms used in our paper are provided by the Torchattacks library~\cite{kim2020torchattacks}.

\noindent\textbf{Alignment in the embedding space:} When using a witness model with a different architecture than the source model, the dimensions of their hidden representations are likely to be different, i.e., $\text{dim}(z_s^{[q_s]}(x)) \neq \text{dim}(z_w^{[q_w]}(x))$, where $q_s$ and $q_w$ represent the layers just before the fully-connected layer in their respective models. To address this dimensional mismatch, we follow previous work~\cite{park2019relational, ma2022distilling, romero2014fitnets, chen2021distilling} and apply a linear projection to $z_s$ so its dimension matches that of $z_w$. The weights of the linear projection are treated as trainable parameters during the alignment process.

\begin{table}[t]
\begin{center}
\renewcommand{\arraystretch}{1.2}
\renewcommand{\tabcolsep}{2.0pt}
\scriptsize
\caption{\textbf{Comparing the similarity between the source and witness model.} We focus on aligning a Res50 with a Res18 and a ViT-B/16 with a ViT-T/16. We first evaluate the similarity between each original source model and its corresponding witness model. This is then followed by an evaluation between the aligned model and the witness model.}
\label{table:similarity}
\begin{tabular}{ ccccccc } 
\Xhline{2\arrayrulewidth}
\multirow{2}{*}{Source/Witness}   & \multicolumn{2}{c}{\underline{KL}} & \multicolumn{2}{c}{\underline{Prediction Agreement}} & \multicolumn{2}{c}{\underline{Input Gradient Cosine Similarity}}\\ 
            & Before & After & Before & After & Before & After\\\Xhline{2\arrayrulewidth}
Res50 / Res18  	    & 0.63    & 0.31   & 80.7\%    & 82.4\% & 0.043    & 0.096     \\ 
ViT-B / ViT-T	    & 0.75    & 0.24   & 78.9\%    & 85.0\% & 0.026    & 0.063    \\
\Xhline{2\arrayrulewidth}
\end{tabular}
\end{center}
\end{table}
\section{Understanding Model Alignment}\label{appendix:understanding}
\subsection{Evaluating Similarity Between the Source and Witness Model}
To study the similarity between the source and witness models before and after alignment, we examine KL divergence, prediction agreement, and cosine similarity of input gradients. The latter is especially important for its role in the attack processes. Our investigation involves aligning a Res50 with a Res18 and a ViT-B/16 with a ViT-T/16. We first evaluate the similarity between each original source model and its corresponding witness model. This is then followed by an evaluation between the aligned model and the witness model. Tab.~\ref{table:similarity} presents the results, indicating reductions in KL divergence, increased cosine similarity, and improved prediction agreement post-alignment. 

\subsection{Smoothness Analysis Using the Largest Eigenvalue of the Input Hessian}\label{appendix:input_hessian}
\begin{table}[t]
\begin{center}
\renewcommand{\arraystretch}{1.1}
\renewcommand{\tabcolsep}{6.0pt}
\scriptsize
\caption{ \textbf{Comparing the largest eigenvalue of the input Hessian on the original source model and the aligned source model.} 
Results are averaged over 1000 randomly selected data points from the ImageNet test set. The largest eigenvalue decreases significantly when evaluated on all three types of inputs.
} 
\label{table:input_hessian_comparison}
\begin{tabular}{ ccc } 
\Xhline{2\arrayrulewidth}
\multirow{2}{*}{Data}  & \multicolumn{2}{c}{\underline{Res50}} \\ 
  & Original & Aligned  \\\Xhline{2\arrayrulewidth}
Clean  	            & 4.16    & 1.25        \\ 
Gaussian-perturbed	& 4.34    & 1.27        \\
PGD-perturbed		& 0.10    & 0.05        \\
\Xhline{2\arrayrulewidth}
\end{tabular}
\end{center}

\end{table}
In Sec.~3.3, we study the effect of model alignment from a geometric perspective. Our analysis demonstrates a smoothing effect on the loss surface due to model alignment, as shown by the significant decrease in gradient norm.

Another metric for evaluating the smoothness of the loss landscape is the largest eigenvalue of the input Hessian~\cite{zhao2020bridging}. This analysis, based on a second-order Taylor expansion of the loss function, assumes that the local curvature of loss with respect to the input can be well represented by the eigenspectrum of the Hessian matrix. Concretely, we measure $\lambda_{\max}(H)$ where $H = \nabla^2_x \ell(x+\Delta x, y, \theta)$, for $\Delta x \in \set{0, \text{Gaussian}, \text{PGD}}$ and for $\theta \in \set{\theta_s, \theta_a}$ over 1000 randomly sampled images from the ImageNet test set.

The results are summarized in Tab.~\ref{table:input_hessian_comparison}. We only incude results for Res50, as PyTorch currently does not support computing second-order derivatives for ViT models. In line with the result presented in Sec.~3.3, the decrease in the largest eigenvalue for all inputs demonstrates a smoothing effect on the loss surface due to the alignment process.

\begin{table}[t]
\begin{center}
\renewcommand{\arraystretch}{1.3}
\renewcommand{\tabcolsep}{2.0pt}
\tiny
\caption{ \textbf{Aligning the Res50 source model with multiple Res18 as witness models.} Using an increasing number of Res18 witness models during the alignment process results in a modest improvement in transferability for adversarial examples generated from the aligned Res50.
This result suggests that simply quadrupling the number of witness models, without considering their diversity, does not lead to a proportional improvement in transferability. }
\label{table:num_witness}
\begin{tabular}{ cllllllllll }
\Xhline{2\arrayrulewidth}
\multirow{2}{*}{\# Witness}	& \multicolumn{10}{c}{Target}\\ \cline{2-11}
 &Res18 & Res50 &	Res101 &	VGG19&	DN121&	IncV3&	ViT-T/16&	ViT-S/16&	ViT-B/16&	SWIN \\ \Xhline{2\arrayrulewidth}

n/a & $44.52$	& $61.51$	& $53.01$	& $43.47$	& $50.59$	& $36.05$	& $25.58$	& $21.54$	& $18.86$	& $22.65$ \\
1 & $+35.30$	& $+26.56$	& $+25.20$	& $+24.38$	& $+27.86$	& $+19.18$	& $+5.49$	& $+2.85$	& $+1.61$	& $+9.60$ \\
2 & $+35.01$	& \gc$+27.83$	& $+24.82$	& $+24.50$	& $+27.88$	& $+19.31$	& $+8.28$	& $+3.82$	& $+1.19$	& $+10.67$ \\
3 & $+34.69$	& $+27.69$	& $+27.39$	& $+24.37$	& \gc$+29.44$	& $+21.51$	& $+8.19$	& $+4.49$	& $+3.27$	& $+11.74$ \\
4 & \gc$+35.57$	& $+27.63$	& \gc$+27.94$	& \gc$+25.12$	& $+28.61$	& \gc$+21.81$	& \gc$+8.29$	& \gc$+5.00$	& \gc$+3.78$	& \gc$+12.15$ \\

\Xhline{2\arrayrulewidth}
\end{tabular}
\end{center}
\end{table}
\section{Additional Experiment}\label{appendix:exp}
\subsection{More Witness Models, Higher Transferability}\label{appendix:exp_num_witness}
Evaluations in Sec.~4 focus on alignment using a single witness model. However, aligning the source model with multiple witness models can encourage learning features extracted by a group of witness models, potentially increasing transferability even further. This approach also serves as an effective strategy to prevent overfitting to a single witness model. Tab.~\ref{table:num_witness} demonstrates the results of using an increasing number of Res18 models to align with Res50. While we do observe that a greater number of witness models tends to result in more transferable perturbations, the improvement is modest, indicating that simply increasing the number of witnesses without considering their diversity does not proportionally enhance transferability. 

\subsection{Witness Model Diversity Matters}\label{appendix:diversity}
Though having more witnesses can prevent overfitting, model diversity is crucial. For example, in Tab.~\ref{table:multi_witness}, we compare aligning a Res50 with a single Res50 to aligning it with a Res50, ViT-S, and IncV3. The results show that aligning with models of different architectures improves transferability to some targets while slightly decreasing it for others. Results in Tab.~\ref{table:num_witness} and ~\ref{table:multi_witness} suggest that a strategy for selecting the optimal number and type of witness models is an interesting direction for future research.
\begin{table}[t]
\caption{\textbf{Aligning Res50 using witness models with different architectures.} When combining multiple models of different architectures as witness models, we observe improvements for some target models.}
\label{table:multi_witness}
\begin{center}
\renewcommand{\arraystretch}{1.2}
\renewcommand{\tabcolsep}{2.4pt}
\scriptsize
\begin{tabular}{ cccccccc } 
\Xhline{2\arrayrulewidth}
\multirow{2}{*}{Witness}  & \multicolumn{7}{c}{Target} \\ \cline{2-8}
                        & Res18	&Res50  &VGG19	&DN121	&IncV3	&ViT-S	&SWIN \\\Xhline{2\arrayrulewidth}
Res50  	                & 54.13	&71.41	&50.06	&59.56	&40.84	&22.19	&25.30     \\ 
Res50+IncV3+ViT-S	    & 51.21	&66.25	&48.65	&\gc{59.82}	&\gc{44.17}	&\gc{23.86}	&\gc{25.55}     \\
\Xhline{2\arrayrulewidth}
\end{tabular}
\end{center}
\end{table}
\begin{table}[t]
\caption{\textbf{Model alignment can be integrated with model modification-based methods.} Using a Res50 as the target model, we compare the transferabilty of MI-FGSM-LinBP and PGD-BPA perturbations from the aligned Res50 to those from the original, unaligned Res50. We observe increased transferability across all target models.}
\label{table:linbp}
\begin{center}
\renewcommand{\arraystretch}{1.2}
\renewcommand{\tabcolsep}{2.4pt}
\scriptsize
\begin{tabular}{ ccccccccc } 
\Xhline{2\arrayrulewidth}
\multirow{2}{*}{Attack} &\multirow{2}{*}{Source}  & \multicolumn{7}{c}{Target} \\ \cline{3-9}
                        &       & Res18	&Res50  &VGG19	&DN121	&IncV3	&ViT-T	&SWIN \\\Xhline{2\arrayrulewidth}
\multirow{2}{*}{MI-FGSM-LinBP}  &Res50  & 87.89	&93.55	&84.08	&88.57	&71.19	&43.16	&46.19     \\ 
                                &Res50*	& \gc{97.27}	&\gc{96.27}	&\gc{93.07}	&\gc{95.61}	&\gc{86.04}	&\gc{56.25}	&\gc{57.71}     \\\hline
\multirow{2}{*}{PGD-BPA}        &Res50  & 51.06	&63.98	&51.54	&58.24	&34.02	&12.0	&11.04     \\ 
                                &Res50*	& \gc{76.76}	&\gc{81.98}	&\gc{70.62}	&\gc{77.98}	&\gc{53.18}	&\gc{19.3}	&\gc{20.06}     \\
\Xhline{2\arrayrulewidth}
\end{tabular}

\end{center}
\end{table}

\subsection{Comparison with Model Modification-based Methods}\label{appendix:model_modification}
A key advantage of our approach over model modification-based methods is its model-agnostic nature: alignment can be applied to any model without changing its forward or backward pass. In contrast, other methods require changes, such as those seen in LinBP and BPA, or even complete retraining from scratch. Additionally, our technique can be integrated with these model modification-based methods. In Tab.~\ref{table:linbp}, we observe the transferability improvement of MI-FGSM-LinBP and PGD-BPA generated using the aligned Res50 across all considered target models. 

Note that the evaluation data selection strategy in Tab.~\ref{table:linbp} differs slightly from that in Sec.~4, where perturbations are generated from both the source and target models, and transferability evaluations are based on those misclassified by their originating models. Our evaluation considers a wide range of target model architectures. However, since the attack algorithms are only avaiable for limited architectures, making it difficult to generate perturbations from some target models. As such, results for MI-FGSM-LinBP and PGD-BPA are evalated on 1000 randomly selected inputs and are not included in Tab.~5 with other attacks.

\subsection{Improved Transferability on Defended Models}\label{appendix:defended_models}
To demonstrate the improved transferability on defended models, we evaluate using a normally trained Res50 with five defense mechanisms, including Bit-Red~\cite{xu2017feature}, JPEG~\cite{guo2018countering}, FD~\cite{liu2019feature}, RS~\cite{cohen2019certified}, NRP~\cite{naseer2020self}. We adhere to the exact configuration of the defense methods as described in the previous work~\cite{wang2021enhancing}. We also consider an adversarially trained Res50 (AT)~\cite{wong2020fast}. Results in Tab.~\ref{table:defense} show that although the defense methods can generally reduce transferability, using the aligned model still results in a significant improvement in transferability compared to the original source model.
\begin{table}[t]
\caption{\textbf{Improved transferability on Res50 with defense methods.} Although the defense methods can generally reduce transferability, using the aligned model still results in a significant improvement in transferability compared to the original model.}
\label{table:defense}
\begin{center}
\renewcommand{\arraystretch}{1.2}
\renewcommand{\tabcolsep}{2.4pt}
\scriptsize
\begin{tabular}{ cccccccc } 
\Xhline{2\arrayrulewidth}
\multirow{2}{*}{Source}  & \multicolumn{7}{c}{Defense} \\  \cline{2-8}
                         & n/a & AT~\cite{wong2020fast} & Bit-Red~\cite{xu2017feature} & JPEG~\cite{guo2018countering} & FD~\cite{liu2019feature} & RS~\cite{cohen2019certified} & NRP~\cite{naseer2020self} \\\Xhline{2\arrayrulewidth}
Res50  	    & 61.51 & 48.06 & 49.22	& 38.81	& 43.77	& 35.41	& 33.01  \\ 
Res50*	    & \gc{88.07} & \gc{52.15}	& \gc{54.70}	& \gc{53.03}	& \gc{50.59}	& \gc{43.99}	& \gc{41.19}  \\
\Xhline{2\arrayrulewidth}
\end{tabular}
\end{center}
\end{table}


\begin{table}[t]
\caption{\textbf{Generalizability of model alignment on other datasets.} We consider two additional datasets: Stanford Cars and Food101. On Stanford Cars, SWIN is aligned using Res50. On Food101, Res50 is aligned using \mbox{ConvNeXt-B}. The aligned source model is denoted with *. We demonstrate that improved transferability can also be observed on other datasets.}
\label{table:dataset}

\begin{center}
\renewcommand{\arraystretch}{1.2}
\renewcommand{\tabcolsep}{2.4pt}
\scriptsize
\begin{tabular}{ cccc } 
\Xhline{2\arrayrulewidth}
\multirow{2}{*}{Dataset} & \multirow{2}{*}{Source}  & \multicolumn{2}{c}{Target} \\ \cline{3-4}
                         & & ConvNeXt-B & ViT-B \\\Xhline{2\arrayrulewidth}
\multirow{2}{*}{Stanford Cars} &SWIN  	    & 68.77    & 17.6      \\ 
                               &SWIN*	     & \gc{88.89}    & \gc{50.67}      \\
\Xhline{2\arrayrulewidth}
\end{tabular}
\quad \quad
\begin{tabular}{ cccc } 
\Xhline{2\arrayrulewidth}
\multirow{2}{*}{Dataset} & \multirow{2}{*}{Source}  & \multicolumn{2}{c}{Target} \\ \cline{3-4}
                         & & SWIN & ViT-B \\\Xhline{2\arrayrulewidth}
\multirow{2}{*}{Food101} &Res50  	    & 26.67    & 12.36      \\ 
                         &Res50*	       & \gc{31.88}    & \gc{16.16}      \\
\Xhline{2\arrayrulewidth}
\end{tabular}
\end{center}

\end{table}
\begin{table}[t]
\begin{center}
\renewcommand{\arraystretch}{1.2}
\renewcommand{\tabcolsep}{2.0pt}
\tiny
\caption{ \textbf{Improving transferability via embedding space alignment.} We focus on ViT-B/16 as the source model and ViT-T/16 as the witness model.}
\label{table:kd_method_vit}
\begin{tabular}{ cccccccccccc } 
\Xhline{2\arrayrulewidth}
\multirow{2}{*}{Method}	& \multicolumn{10}{c}{Target}\\ \cline{2-12}
& Res18 &	Res50 &	Res101 &	VGG19&	DN121&	IncV3&	ViT-T/16&	ViT-S/16&	ViT-B/16&	SWIN \\ \Xhline{2\arrayrulewidth}

n/a 								& $23.05$	& $19.67$	& $18.30$	& $21.83$	& $21.96$	& $21.52$	& $36.44$	& $44.58$	& $51.27$	& $21.22$ \\
KL 								& $+10.13$	& $+9.71$	& $+8.42$	& $+10.71$	& $+11.30$	& $+9.89$	& $+51.72$	& $+44.99$	& $+34.69$	& $+20.01$ \\
KD~\cite{park2019relational} 	& \gc$+13.40$	& $+8.47$	& $+8.79$	& $+11.54$	& $+10.25$	& $+11.37$	& $+46.07$	& $+43.41$	& $+34.02$	& $+17.51$ \\
EGA~\cite{ma2022distilling} 		& $+12.55$	& \gc$+10.21$	& $+9.21$	& $+11.39$	& $+9.30$	& $+12.05$	& $+53.08$	& $+44.65$	& $+33.50$	& $+21.40$ \\
HINT~\cite{romero2014fitnets} 	& $+12.78$	& $+10.02$	& \gc$+9.84$	& \gc$+12.21$	& \gc$+12.07$	& \gc$+12.31$	& \gc$+53.41$	& \gc$+47.05$	& \gc$+38.50$	& \gc$+25.51$ \\
NCE~\cite{chen2021distilling} 	& $+8.90$	& $+8.48$	& $+7.16$	& $+9.00$	& $+7.26$	& $+7.92$	& $+45.09$	& $+41.84$	& $+32.57$	& $+15.47$ \\
\Xhline{2\arrayrulewidth}
\end{tabular}
\end{center}
\end{table}
    
\subsection{Experiments on Additional Datasets}\label{appendix:additional_datasets}
Evaluations in Sec.~4 focuses on the ImageNet dataset. Here, we supplement our results with experiments on Stanford Cars~\cite{krause20133d} and Food101~\cite{bossard2014food}. On Stanford Cars, SWIN is aligned using Res50. On Food101, Res50 is aligned using \mbox{ConvNeXt-B}. Results in Tab.~\ref{table:dataset} show that the improved transferability achieved through model alignment is also evident on other datasets.

\subsection{Embedding-space Alignment on Vision Transformers}\label{appendix:kd_method_vit}
The ablation study in Sec.~4 shows that Res50 generates more transferable perturbations when aligned with Res18 in the embedding space. Results in Tab.~\ref{table:kd_method_vit} show that ViT can also benefit from embedding-space alignment.

 
\else \fi

\end{document}